\documentclass{article}

\usepackage{algorithmic}
\usepackage{algorithm}

\usepackage[final,nohyperref]{corl_2017}

\usepackage{amsmath}
\usepackage{amsfonts}
\usepackage{mathtools}
\usepackage{multirow}

\usepackage[export]{adjustbox}
\usepackage{enumitem}

\usepackage{wrapfig}

\usepackage{color}

\begin{document}

\title{\mbox{One-Shot Visual Imitation Learning via Meta-Learning}}

\newcommand\blfootnote[1]{%
  \begingroup
  \renewcommand\thefootnote{}\footnote{#1}%
  \addtocounter{footnote}{-1}%
  \endgroup
}

\author{
  Chelsea Finn*$^1$, Tianhe Yu*$^1$, Tianhao Zhang$^1$, Pieter Abbeel$^{1,2}$, Sergey Levine$^1$\\
  $^1$University of California, Berkeley, 
  $^2$OpenAI 
  \\
  \texttt{cbfinn,tianhe.yu,tianhao.z,pabbeel,svlevine@berkeley.edu} \\
}



%

\maketitle

\begin{abstract}
\blfootnote{*denotes equal contribution}
In order for a robot to be a generalist that can perform a wide range of jobs, it must be able to acquire a wide variety of skills quickly and efficiently in complex unstructured environments. High-capacity models such as deep neural networks can enable a robot to represent complex skills, but learning each skill from scratch then becomes infeasible.
In this work, we present a meta-imitation learning method that enables a robot to learn how to learn more efficiently, allowing it to acquire new skills from just a single demonstration.
Unlike prior methods for one-shot imitation, our method can scale to raw pixel inputs and requires data from significantly fewer prior tasks for effective learning of new skills. Our experiments on both simulated and real robot platforms demonstrate the ability to learn new tasks, end-to-end, from a single visual demonstration.
\end{abstract}


\section{Introduction}

Enabling robots to be generalists, capable of performing a wide variety of tasks with many objects, presents a major challenge for current methods. Learning-based approaches offer the promise of a generic algorithm for acquiring a wide range of skills. However, learning methods typically require a fair amount of supervision or experience per task, especially for learning complex skills from raw sensory inputs using deep neural network models. Moreover, most methods provide no mechanism for using experience from previous tasks to more quickly solve new tasks.
Thus, to learn many skills, training data would need to be collected independently for each and every task.
By reusing data across skills, robots should be able to amortize their experience
and significantly improve data efficiency, requiring minimal supervision for each new skill. In this paper, we consider the question: how can we leverage information from previous skills to quickly learn new behaviors?

\begin{wrapfigure}{r}{0.33\textwidth}
    \begin{center}
    \vspace{-0.6cm}
        \includegraphics[width=0.33\textwidth]{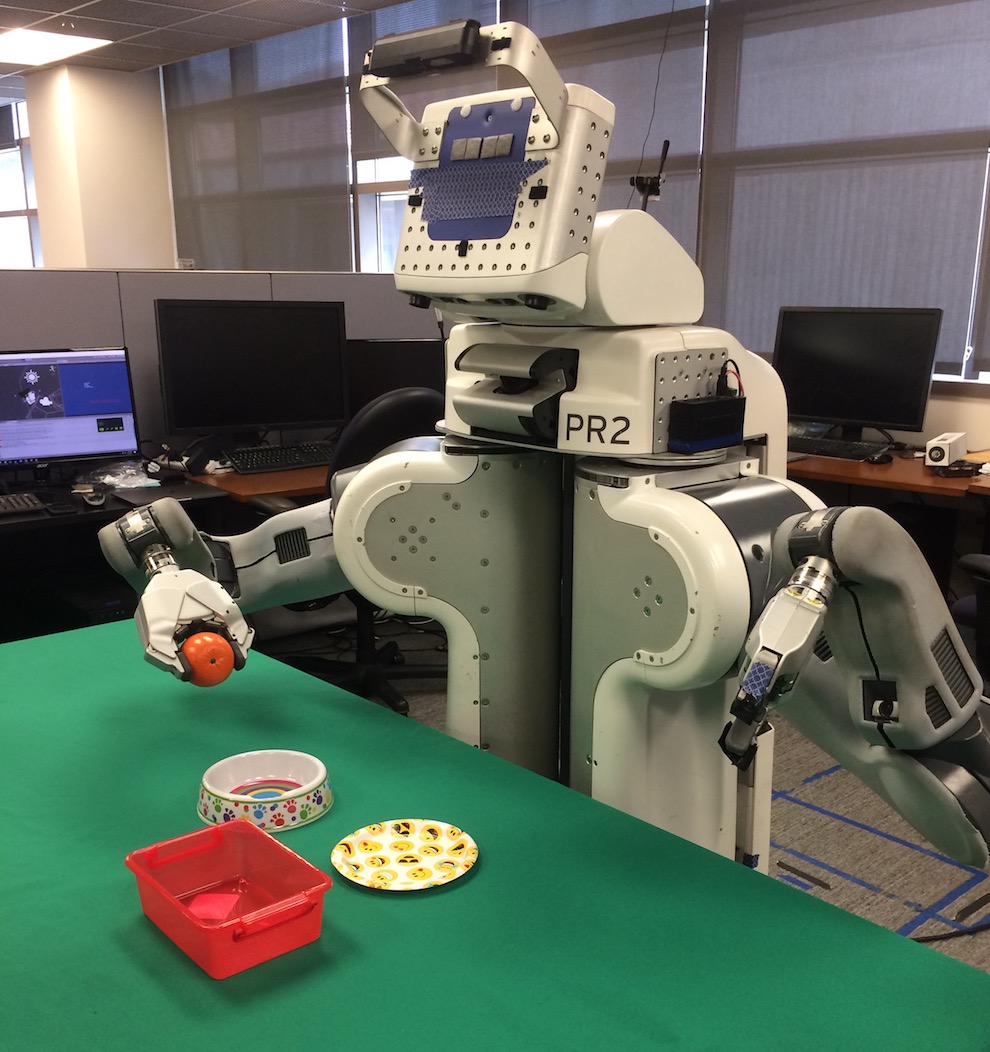}
    \end{center}
    \vspace{-0.3cm}
    \caption{\small The robot learns to place a new object into a new container from a single demonstration.}
    \vspace{-0.5cm}
    \label{fig:teaser}
\end{wrapfigure}

We propose to combine meta-learning with imitation, enabling a robot to reuse past experience and, as a result, learn new skills from a single demonstration.
Unlike prior methods that take the task identity~\cite{multitaskpolicy,contextualpolicy,uvfa,parameterizedskills}
or a demonstration~\cite{rocky} as the input into a contextual policy,
our approach learns a parameterized policy that can be adapted to different tasks through gradient updates, effectively learning to imitation learn.
As a result, the set of skills that can be learned is more flexible while using fewer overall parameters. For the first time, we demonstrate that vision-based policies can be fine-tuned end-to-end from one demonstration,
using meta-learning as a pre-training procedure that uses demonstrations on a diverse range of other environments.

The primary contribution of this paper is to demonstrate an approach for one-shot imitation learning from raw pixels.
We evaluate our approach on two simulated planar reaching domains, on simulated pushing tasks, and on visual placing tasks on a real robot (See Figure~\ref{fig:teaser}).
Our approach is able to learn visuomotor policies that can adapt to new task variants using only one visual demonstration,  including settings where only a raw video of the demonstration is available without access to the controls applied by the demonstrator. 
By employing a parameter-efficient meta-learning method, our approach requires a relatively modest number of demonstrations for meta-learning and scales to raw pixel inputs. 
As a result, our method can successfully be applied to real robotic systems.

\section{Related Work}

We present a method that combines imitation learning~\cite{imitation_survey} with meta-learning~\cite{thrun} for one-shot learning from visual demonstrations. Efficient imitation from a small number of demonstrations has been successful in scenarios where the state of the environment, such as the poses of objects, is known~\cite{billard,movement_primitives,pastor,ratliff}.
In this work, we focus on settings where the state of the environment is unknown, where we must instead learn from raw sensory inputs. This removes the need for pre-defined vision systems while also making the method applicable to vision-based non-prehensile manipulation
in unknown, dynamic environments.
Imitation learning from raw pixels has been widely studied in the context of mobile robotics~\cite{alvinn,nvidia,idsia,safedagger}.
However, learning from demonstrations has two primary challenges when applied to real-world settings. The first is the widely-studied issue of compounding errors~\cite{dagger,shiv_laskey,safedagger}, which we do not address in this paper. The second is the need for a large number of demonstrations for each task. This latter limitation is a major roadblock for developing generalist robots that can learn a wide variety of tasks through imitation. Inverse reinforcement learning~\cite{ngrussell} can reduce the number of demonstrations needed by inferring the reward function underlying a set of demonstrations.
However, this requires additional robot experience to optimize the reward~\cite{gcl,ropes,pierre}. This experience typically comes in the form of trial-and-error learning or data for learning a model.

In this work, we drastically reduce the number of demonstrations needed for an individual task by sharing data across tasks. In particular, our goal is to learn a new task from a single demonstration of that task by using a dataset of demonstrations of many other tasks for meta-learning.
Sharing information across tasks is by no means a new idea, e.g. by using task-to-task mappings~\cite{transfer}, gating~\cite{tennismotions}, and shared features~\cite{invariantfeatures}. These multi-task robotic learning methods  consider the problem of generalization to new tasks from some specification of that task.
A common approach, often referred to as contextual policies, is to provide the task as an input to the policy or value function, where the task is represented as a goal or demonstration~\cite{multitaskpolicy,contextualpolicy,parameterizedskills,uvfa,rocky}. 
Another approach is to train controllers for a variety of tasks and learn a mapping from task representations to controller parameters~\cite{pastor,adjustdmp,parametrizedskills}.
In this work, we instead use meta-learning to enable the robot to quickly learn new tasks with gradient-based policy updates. 
In essence, we learn policy parameters that, when finetuned on just one demonstration of a new task, can immediately learn to perform that task.
This enables the robot to learn new tasks end-to-end with extreme efficiency, using only one demonstration, without requiring any additional mechanisms such as contexts or learned update functions.

\newcommand{\task}{\mathcal{T}}
\newcommand{\loss}{\mathcal{L}}
\newcommand{\inp}{\mathbf{o}}
\newcommand{\out}{\mathbf{a}}
\newcommand{\learner}{f}
\newcommand{\policy}{\pi}
\newcommand{\lossi}{\loss_{\task_i}}
\newcommand{\losst}{\loss_{\task}}

\section{Meta-Imitation Learning Problem Formulation}

In this section, we introduce the visual meta-imitation learning problem, where a vision-based policy must adapt to a new task from a single demonstration. We also summarize a prior meta-learning method that we will extend into a meta-imitation learning algorithm in Section~\ref{sec:mil}.

\subsection{Problem Statement}
\label{sec:problem}

Our goal is to learn a policy that can quickly adapt to new tasks from a single demonstration of that task. To remove the need for a large amount of task-specific demonstration data, we propose to reuse demonstration data from a number of other tasks to enable efficient learning of new tasks. By training for adaptation across tasks, meta-learning effectively treats entire tasks as datapoints. The amount of data available for each individual task is relatively small. In the context of robotics, this is precisely what we want for developing generalist robots -- the ability to provide a small amount of supervision for each new task that the robot should perform. In this section, we will formally define the one-shot imitation learning problem statement and introduce notation.

We consider a policy $\policy$ that maps observations $\inp$ to predicted actions $\hat{\out}$.
During meta-learning, the policy
is trained to adapt to a large number of tasks.
Formally, each imitation task $\task_i = \{ \tau = \{\inp_1,\out_1,\dots,\inp_T,\out_T\} \sim \policy_i^\star, \loss(\out_{1:T}, \hat{\out}_{1:T}), T  \}$
consists of demonstration data $\tau$ generated by an expert policy $\pi_i^\star$ and
a  loss function $\loss$ used for imitation. 
Feedback is provided by the loss function $\loss(\out_1, ..., \out_T, \hat{\out_1}, ..., \hat{\out_T})\rightarrow \mathbb{R}$, which might be mean squared error for continuous actions, or a cross-entropy loss for discrete actions. 

In our meta-learning scenario, we consider a distribution over tasks $p(\task)$.
In the one-shot learning setting, the policy is trained to learn a new task $\task_i$ drawn from $p(\task)$ from only one demonstration  generated by $\task_i$.
During meta-training, a task $\task_i$ is sampled from $p(\task)$, the policy is trained using one demonstration from an expert $\pi_i^\star$ on $\task_i$, and then tested on a new demonstration from $\pi_i^\star$ to determine its training and test error according to the loss $\loss$.
The policy $\policy$ is then improved by considering how the \emph{test} error on the new demonstration changes with respect to the parameters. Thus, the test error on sampled demonstration from $\pi_i^\star$ serves as the training error of the meta-learning process.
At the end of meta-training, new tasks are sampled from $p(\task)$,
and meta-performance is measured by the policy's performance after learning from one demonstration.
Tasks used for meta-testing are held out during meta-training.

\subsection{Background: Model-Agnostic Meta-Learning}
\label{sec:maml}

In our approach to visual meta-imitation learning, we will use meta-learning to train for fast adaptation across a number of tasks by extending model-agnostic meta-learning (MAML)~\cite{maml} to meta-imitation learning from visual inputs. Previously, MAML has been applied to few-shot image recognition and reinforcement learning.
The MAML algorithm aims to learn the weights $\theta$ of a model $\learner_\theta$ such that standard gradient descent can make rapid progress on new tasks $\task$ drawn from $p(\task)$, without overfitting to a small number of examples.
Because the method uses gradient descent as the optimizer, it does not introduce any additional parameters, making it more parameter-efficient than other meta-learning methods.
When adapting to a new task $\task_i$, the model's parameters $\theta$ become $\theta_i'$. In MAML, the updated parameter vector $\theta_i'$ is computed using one or more gradient descent updates on task $\task_i$, i.e. $\theta_i'=\theta-\alpha \nabla_\theta  \lossi(  \learner_\theta )$. For simplicity of notation, we will consider one gradient update for the rest of this section, but using multiple gradient updates is a straightforward extension.

The model parameters are trained by optimizing for the performance of $\learner_{\theta_i'}$ with respect to $\theta$ across tasks sampled from $p(\task)$, corresponding to the following problem:
\begin{align}
\label{eq:meta}
\min_\theta \sum_{\task_i \sim p(\task)}  \lossi ( \learner_{\theta_i'}) 
= \sum_{\task_i \sim p(\task)}  \lossi ( \learner_{\theta - \alpha \nabla_\theta \lossi(f_\theta)})
\end{align}
Note that the meta-optimization is performed over the parameters $\theta$, whereas the objective is computed using the updated parameters $\theta'$. In effect, MAML aims to optimize the model parameters such that one or a small number of gradient steps on a new task will produce maximally effective behavior on that task. The meta-optimization across tasks uses stochastic gradient descent (SGD). 

\vspace{-0.1cm}
\section{Meta-Imitation Learning with MAML} 
\label{sec:mil}
\vspace{-0.1cm}

\begin{algorithm}[t]
\caption{Meta-Imitation Learning with MAML}
\label{alg:main}
\begin{algorithmic}[1]
{\footnotesize
\REQUIRE $p(\task)$: distribution over tasks
\REQUIRE $\alpha$, $\beta$: step size hyperparameters
\STATE randomly initialize $\theta$
\WHILE{not done}
\STATE Sample batch of tasks $\task_i \sim p(\task)$
  \FORALL{$\task_i$}
      \STATE Sample demonstration $\tau=\{\inp_1, \out_1, ... \inp_T, \out_T\}$ from $\task_i$
      \STATE Evaluate $\nabla_\theta \lossi(\learner_\theta)$ using $\tau$ and $\lossi$ in Equation~(\ref{eq:sup1})
      \STATE Compute adapted parameters with gradient descent: $\theta_i'=\theta-\alpha \nabla_\theta  \lossi(  \learner_\theta )$
      \STATE Sample demonstration $\tau_i'=\{\inp_1', \out_1', ... \inp_T', \out_T'\}$ from $\task_i$ for the meta-update
 \ENDFOR
 \STATE Update $\theta \leftarrow \theta - \beta \nabla_\theta \sum_{\task_i \sim p(\task)}  \lossi ( \learner_{\theta_i'})$ using each $\tau_i'$ and $\lossi$ in Equation~\ref{eq:sup1}
\ENDWHILE
\RETURN parameters $\theta$ that can be quickly adapted to new tasks through imitation.
}
\end{algorithmic}
\end{algorithm}

In this section, we describe how we can extend the model-agnostic meta-learning algorithm (MAML) to the imitation learning setting. The model's input, $\inp_t$, is the agent's observation at time $t$, e.g. an image, whereas the output $\out_t$ is the action taken at time $t$, e.g. torques applied to the robot's joints. We will denote a demonstration trajectory as $\tau := \{\inp_1, \out_1, ... \inp_T, \out_T\}$ and use a mean squared error loss as a function of policy parameters $\phi$ as follows:
\vspace{-0.06cm}
\begin{align}
\label{eq:sup1}
\lossi(\learner_\phi ) =  \sum_{\tau^{(j)} \sim \task_i} \sum_{t}   ~ \lVert \learner_\phi(\inp_t^{(j)}) - \out_t^{(j)}  \rVert_2^2.
\end{align}
\vspace{-0.45cm}

We will primarily consider the one-shot case, where only a single demonstration $\tau^{(j)}$ is used for the gradient update. However, we can also use multiple demonstrations to resolve ambiguity.

For meta-learning, we assume a dataset of demonstrations with at least two demonstrations per task. This data is only used during meta-training; meta-test time assumes only one demonstration for each new task.
During meta-training, each meta-optimization step entails the following: A batch of tasks is sampled and two demonstrations are sampled per task. Using one of the demonstrations, $\theta_i'$ is computed for each task $\task_i$ using gradient descent with Equation~\ref{eq:sup1}. Then, the second demonstration of each task is used to compute the gradient of the meta-objective by using Equation~\ref{eq:meta} with the loss in Equation~\ref{eq:sup1}. Finally, $\theta$ is updated according to the gradient of the meta-objective. In effect, the pair of demonstrations serves as a training-validation pair. The algorithm is summarized in Algorithm~\ref{alg:main}.

The result of meta-training is a policy that can be adapted to new tasks using a single demonstration. Thus, at meta-test time, a new task $\task$ is sampled, one demonstration for that task is provided, and the model is updated to acquire a policy for that task. During meta-test time, a new task might involve new goals or manipulating new, previously unseen objects.

\vspace{-0.16cm}
\subsection{Two-Head Architecture: Meta-Learning a Loss for Fast Adaptation}
\vspace{-0.16cm}

In the standard MAML setup, outlined previously,
the policy is consistent across the pre- and post-gradient update stages. However, we can make a modification such that the parameters of the final layers of the network are not shared, forming two ``heads,'' as shown in Figure~\ref{fig:latenttaskvar}. The parameters of the pre-update head are not used for the final, post-update policy, and the parameters of the post-update head are not updated using the demonstration. But, both sets of parameters are meta-learned for effective performance after adaptation. Interestingly, this two head architecture amounts to using a different inner objective in the meta-optimization, while keeping the same outer objective. To see this, let us denote $\mathbf{y}_t^{(j)}$ as the set of post-synamptic activations of the last hidden layer, and $W$ and $b$ as the weight matrix and bias of the final layer. The inner loss function is then given by:
\vspace{-0.06cm}
\begin{align}
\label{eq:sup_act}
\lossi^*(\learner_\phi ) =  \sum_{\tau^{(j)} \sim \task_i} \sum_{t}   ~ \lVert 
 W \mathbf{y}_t^{(j)}+b - \mathbf{a}_t^{(j)}
\rVert_2^2,
\end{align}
\vspace{-0.45cm}

where $W$ and $b$, the weights and bias of the last layer, effectively form the parameters of the meta-learned loss function.
We use the meta-learned loss function $\lossi^*$ to compute the adapted parameter
$\theta_i'$ of each task $\task_i$, via gradient descent.
Then, the meta-objective becomes:
\vspace{-0.1cm}
\begin{align}
\label{eq:meta_two_head}
\min_{\theta, W, b} \sum_{\task_i \sim p(\task)}  \lossi ( \learner_{\theta_i'}) 
= \sum_{\task_i \sim p(\task)}  \lossi ( \learner_{\theta - \alpha \nabla_\theta \lossi^*(f_\theta)}).
\end{align}
\vspace{-0.4cm}

This provides the algorithm more flexibility in how it adapts the policy parameters to the expert demonstration, which we found to increase performance in a few experiments (see Appendix~\ref{app:place}). However, the more interesting implication of using a learned loss is that we can omit the actions during 1-shot adaptation, as we discuss next.

\vspace{-0.16cm}
\subsection{Learning to Imitate without Expert Actions}
\label{sec:no_action}\vspace{-0.16cm}

Conventionally, a demonstration trajectory consists of pairs of observations and actions, as we discussed in Section~\ref{sec:mil}. However, in many scenarios, it is more practical to simply provide a video of the task being performed, e.g. by a human or another robot.
One step towards this goal, which we consider in this paper, is to remove the need for the robot arm trajectory and actions at test time.\footnote{We leave the problem of domain shift, i.e.\! between a video of a human and the robot's view, to future work.}  Though, to be clear, we will assume access to expert actions during meta-training.
Without access to expert actions at test time, it is unclear what the  loss function for 1-shot adaptation should be. Thus, we will meta-learn a loss function, as discussed in the previous section. We can simply modify the loss in Equation~\ref{eq:sup_act} by removing the expert actions:
\vspace{-0.1cm}
$$
\lossi^*(\learner_\phi ) =  \sum_{\tau^{(j)} \sim \task_i} \sum_{t}   ~ \lVert 
 W \mathbf{y}_t^{(j)}+b
\rVert_2^2,
$$
\vspace{-0.5cm}

This corresponds to a learned quadratic loss function on the final layer of activations, with parameters $W$ and $b$.
Though, in practice, the loss function could be more complex. With this loss function, we can learn to learn from the raw observations of a demonstration using the meta-optimization objective in Equation~\ref{eq:meta_two_head}, as shown in our experiments in Sections~\ref{sec:pushing} and~\ref{sec:placing}.

\vspace{-0.1cm}
\section{Model Architectures for Meta-Imitation Learning}
\label{sec:arch}
\vspace{-0.2cm}

\begin{figure}[t!]
\setlength{\unitlength}{0.5\columnwidth}
\begin{picture}(1.99,0.74) \linethickness{0.5pt}
\put(0.1,0.4){\includegraphics[width=0.9\columnwidth]{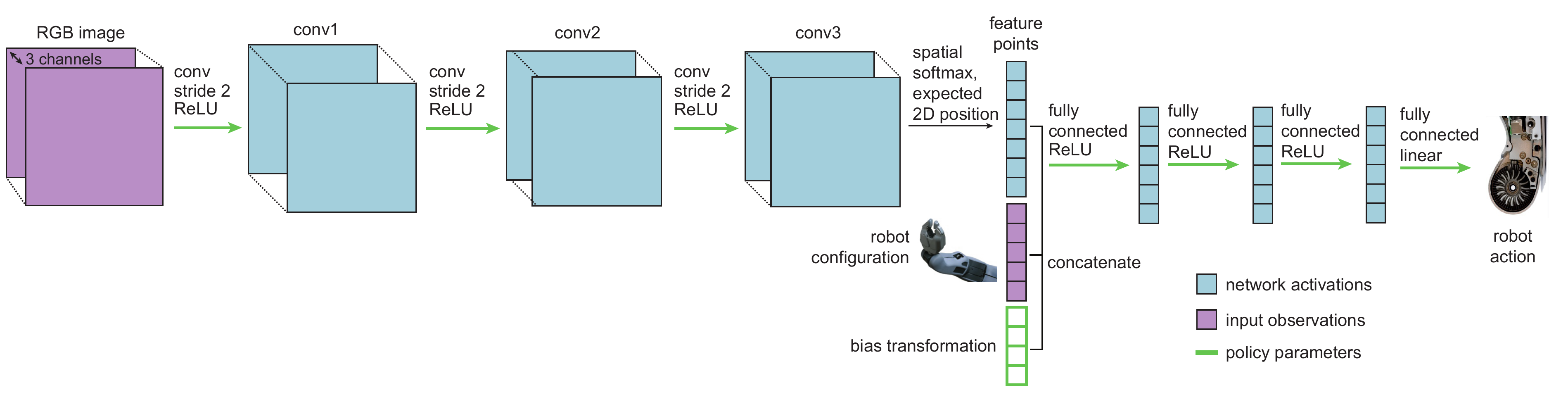}}
\put(0.1,-0.05){\includegraphics[width=0.9\columnwidth]{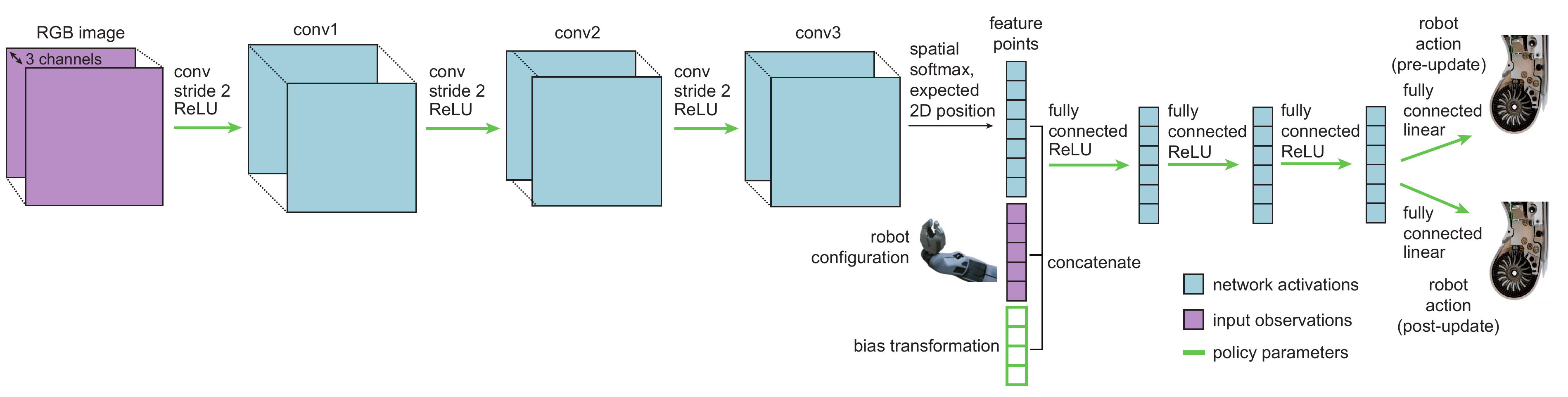}}

\end{picture}
\vspace{-0.1cm}
\caption{\footnotesize Diagrams of the policy architecture with a bias transformation (top and bottom) and two heads (bottom). The green arrows and boxes indicate weights that are part of the meta-learned policy parameters $\theta$.
\label{fig:latenttaskvar} 
\vspace{-0.4cm}
}
\end{figure}

We use a convolutional neural network (CNN) to represent the policy, similar to prior vision-based imitation and meta-learning methods~\cite{nvidia,maml}. The policy observation includes both the camera image and the robot's configuration, e.g. the joint angles and end-effector pose. 
In this section, we overview the policy architecture, but leave the details to be discussed in Section~\ref{sec:experiments}.
The policy consists of several strided convolutions, followed by ReLU nonlinearities.
The final convolutional layer is transformed into spatial feature points using a spatial soft-argmax~\cite{e2e,dsae} and concatenated with the robot's configuration. The result is passed through a set of fully-connected layers with ReLU nonlinearities. Because the data within a demonstration trajectory is highly correlated across time,
batch normalization was not effective. Instead, we used layer normalization after each layer~\cite{layernorm}.

Although meta-imitation learning can work well with standard policy architectures such as the one described above, the optimal architecture for meta-learning does not necessarily correspond to the optimal architecture for standard supervised imitation learning. One particular modification that we found improves meta-learning performance is to concatenate a vector of parameters
 to a hidden layer of post-synaptic activations, which leads to what we will refer to as a \emph{bias transformation}. This parameter vector is treated the same as other parameters in the policy during both meta-learning and test-time adaptation.
Formally, let us denote the parameter vector as $\mathbf{z}$, the post-synaptic activations
as $\mathbf{x}$, and the pre-synaptic activations at the next layer as $\mathbf{y}$. A standard neural network architecture sets $\mathbf{y} = W\mathbf{x} + b$, for bias $b$ and weight matrix $W$.
The error gradient with respect to the standard bias $\frac{d \loss}{d b}$ equals the error gradient with respect to $y$, $\frac{d \loss}{d y}$. Thus, a gradient update of the standard bias is directly coupled with the update to the weight matrix $W$ and parameters in earlier layers of the network.
The bias transformation, which we describe next, provides more control over the updated bias by eliminating this decoupling. With a bias transformation, we set $\mathbf{y} = W_1 \mathbf{x} + W_2 \mathbf{z} + b$,
where $W =[W_1, W_2]$ and $b$ are the weight matrix and bias. First, note that including $\mathbf{z}$ and $W_2$ simply corresponds to a reparameterization of the bias, $\tilde{b} = W_2 \mathbf{z} + b$, since neither $W_2 \mathbf{z}$ nor $b$ depend on the input. The error gradient with respect to $\mathbf{z}$ and $W_2$ are: $\frac{d \loss}{d W_2}  = \frac{d \loss}{d y} \mathbf{z}^T $ and $\frac{d \loss}{d \mathbf{z}}  = W_2^T \frac{d \loss}{d y} $. 
After one gradient step, the updated transformed bias is: $\tilde{b}' = (W_2 - \alpha  \frac{d \loss}{d y} \mathbf{z}^T) (\mathbf{z} - \alpha W_2^T \frac{d \loss}{d y})+b - \alpha \frac{d \loss}{d y}$.
Thus, a gradient update to the transformed bias can be controlled more directly by the values of $W_2$ and $\mathbf{z}$, whose values do not directly affect the gradients of other parameters in the network.
To see one way in which the network might choose to control the bias, consider the setting where $\mathbf{z}$ and $b$ are zero. Then, the updated bias is:
$\tilde{b}' =-\alpha  W_2  W_2^T\frac{d \loss}{d y} -\alpha \frac{d \loss}{d y}$. In summary, the bias transformation increases the representational power of the \emph{gradient}, without affecting the representation power of the network itself.
In our experiments, we found this simple addition to the network made gradient-based meta-learning significantly more stable and effective.
We include a diagram of  the policy architecture with the bias transformation in Figure~\ref{fig:latenttaskvar}.

\vspace{-0.1cm}
\section{Experiments}
\label{sec:experiments}
\vspace{-0.2cm}

The goals of our experimental evaluation are to answer the following questions: (1) can our method learn to learn a policy that maps from image pixels to actions using a single demonstration of a task? (2) how does our meta-imitation learning method compare to prior one-shot imitation learning methods with varying dataset sizes? (3) can we learn to learn without expert actions? (4) how well does our approach scale to real-world robotic tasks with real images?

We evaluate our method on one-shot imitation in three experimental domains. In each setting, we compare our proposed method to a subset of the following methods:
\vspace{-0.2cm}
\begin{itemize}[leftmargin=*]
\item \textbf{random policy}: A policy that outputs random actions from a standard Normal distribution. 
\item \textbf{contextual policy}: A feedforward policy, which takes as input the final image of the demonstration, to indicate the goal of the task, and the current image, and outputs the current action.
\item \textbf{LSTM}: A recurrent neural network which ingests the provided demonstration and the current observation, and outputs the current action, as proposed by~\citet{rocky}.
\item \textbf{LSTM+attention}: A recurrent neural network using the attention architecture proposed by~\citet{rocky}. This method is only applicable to non-vision tasks.
\end{itemize}
\vspace{-0.2cm}
The contextual policy, the LSTM policies, and the proposed approach are all trained using the same datasets, with the same supervision. All policies, including the proposed approach, were meta-trained via a behavioral cloning objective (mean squared error) with supervision from the expert actions, using the Adam optimizer with default hyperparameters~\cite{adam}.

\begin{figure}
\setlength{\unitlength}{0.5\columnwidth}
\centering

\includegraphics[width=0.168\linewidth,valign=m]{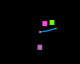}
\includegraphics[width=0.168\linewidth,valign=m]{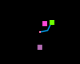}
\includegraphics[width=0.135\linewidth,valign=m]{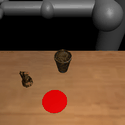}
\includegraphics[width=0.135\linewidth,valign=m]{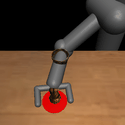}
\includegraphics[width=0.18\linewidth,valign=m]{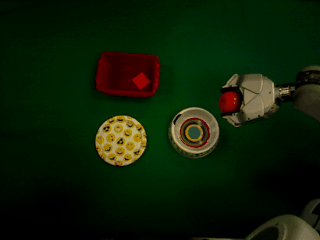}
\includegraphics[width=0.18\linewidth,valign=m]{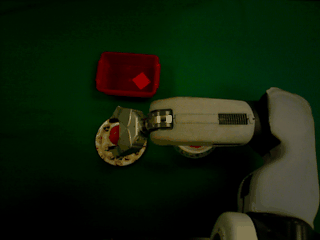}\\
\includegraphics[width=0.168\linewidth,valign=m]{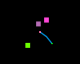}
\includegraphics[width=0.168\linewidth,valign=m]{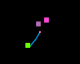} 
\includegraphics[width=0.135\linewidth,valign=m]{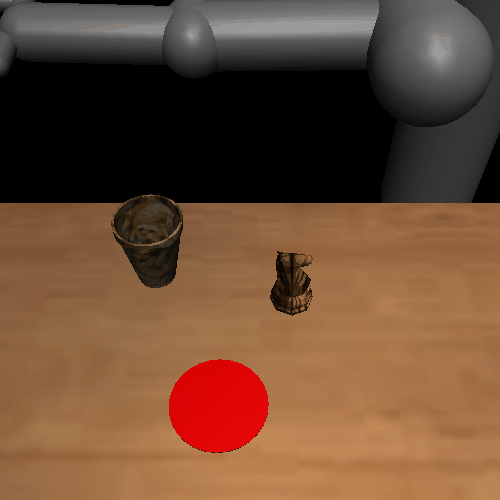}
\includegraphics[width=0.135\linewidth,valign=m]{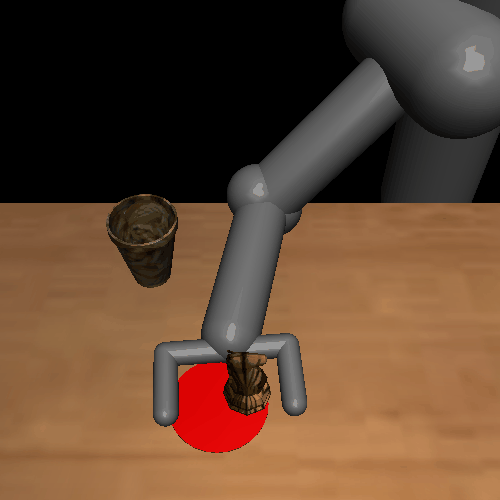} 
\includegraphics[width=0.18\linewidth,valign=m]{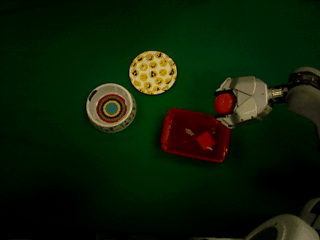}
\includegraphics[width=0.18\linewidth,valign=m]{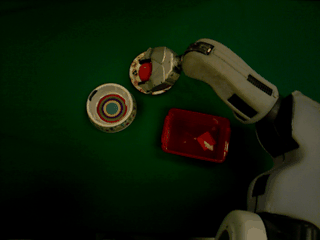}
\vspace{-0.1cm}
\caption{\footnotesize Example tasks from the policy's perspective. In the top row, each pair of images shows the start and final scenes of the demonstration. The bottom row shows the corresponding scenes of the learned policy roll-out.
Left: Given one demonstration of reaching a target of a particular color, the policy must learn to reach for the same color in a new setting. Center: The robot pushes the target object to the goal after seeing a demonstration of pushing the same object toward the goal in a different scene. Right: We provide a demonstration of placing an object on the target, then the robot must place the object on the same target in a new setting.
\label{fig:task}
\vspace{-0.3cm}
}
\end{figure}

\subsection{Simulated Reaching}
\vspace{-0.1cm}

The first experimental domain is a family of planar reaching tasks, as illustrated in Figure~\ref{fig:task}, where the goal of a particular task is to reach a target of a particular color, amid two distractors with different colors. This simulated domain allows us to rigorously evaluate our method and compare with prior approaches and baselines. We consider both vision and non-vision variants of this task, so that we can directly compare to prior methods that are not applicable to vision-based policies. See Appendix~\ref{app:reach} for more details about the experimental setup and choices of hyperparameters.

\begin{figure}
\setlength{\unitlength}{0.5\columnwidth}
\vspace*{1em}
\begin{picture}(1.99,0.32) \linethickness{0.5pt}
\put(1.32,-0.05){
\includegraphics[width=0.34\columnwidth, clip,trim={1.5em 0 2.5em 0}]
{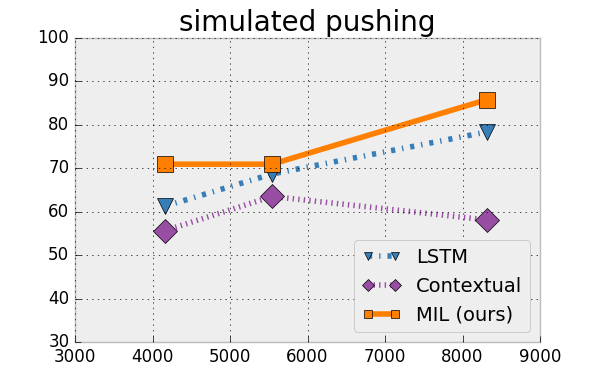}
}
\put(0.66,-0.05){
\includegraphics[width=0.34\columnwidth, clip,trim={1.5em 0 2.5em 0}]
{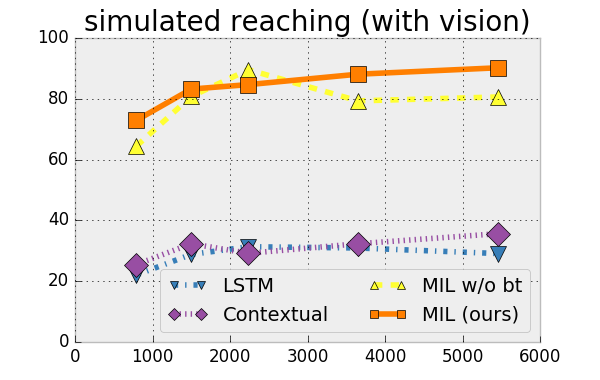}
}
\put(0,-0.05){
\includegraphics[width=0.34\columnwidth, clip,trim={1.5em 0 2.5em 0}]
{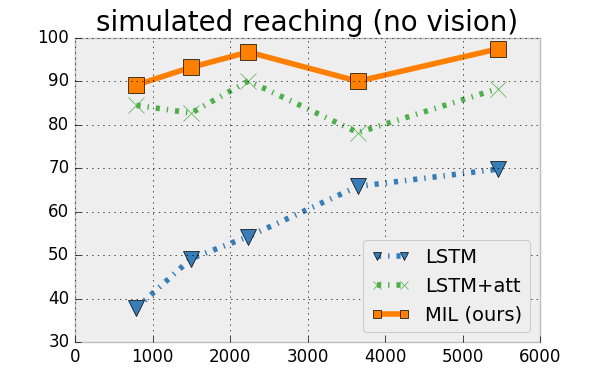}
}
\put(0.47, -0.075){total number of demonstrations in the meta-training set}
\put(0, 0.03){\rotatebox{90}{success rate (\%)}}
\end{picture}
\vspace*{1em}
\caption{\footnotesize
One-shot success rate on test tasks as a function of the meta-learning dataset size in the simulated domains. Our meta-imitation learning approach (MIL) can perform well across a range of dataset sizes, and can more effectively learn new tasks than prior approaches that feed in the goal image (contextual) or demonstration (LSTM) as input. A random policy achieves $25.7\%$ reaching success and $0.45\%$ pushing success. For videos of the policies, see the supplementary video\protect\footnotemark.
}
\label{fig:results-simreach}
\vspace{-0.2cm}
\end{figure}

We evaluate each method on a range of meta-training dataset sizes and  show the one-shot imitation success rate in Figure~\ref{fig:results-simreach}. Using vision, we find that meta-imitation learning is able to handle raw pixel inputs, while the LSTM and contextual policies struggle to learn new tasks using modestly-sized meta-learning datasets. In the non-vision case, which involves far fewer parameters, the LSTM policies fare much better, particularly when using attention, but still perform worse than MIL. Prior work demonstrated these approaches using 10,000 or more demonstrations~\cite{rocky}. Therefore, the mediocre performance of these methods on much smaller datasets is not surprising.
We also provide a comparison with and without the bias transformation discussed in Section~\ref{sec:arch}. The results demonstrate that MIL with the bias transformation (bt) can perform more consistently across dataset sizes.

\vspace{-0.1cm}

\subsection{Simulated Pushing}
\label{sec:pushing}

\vspace{-0.1cm}

\footnotetext{For video results, see \url{https://sites.google.com/view/one-shot-imitation}}

\begin{wraptable}{r}{7.0cm}
\vspace{-1.8cm}
    \begin{center}
    \begin{tabular}{l|c|c|c|c}
    \hline
    \!\!method & & \!\!\! \begin{tabular}{@{}c@{}}video+state \\ +action\end{tabular} \!\!\! &  \!\!\!\begin{tabular}{@{}c@{}}video\\+state  \end{tabular}\!\!\! & video\\
      \hline
      \!\!LSTM &\multirow{3}{*}{\!\!\rotatebox{90}{\!1-shot}\!\!} & 78.38\% &\!\! 37.61\%\!\! & \!\!34.23\%\!\!\\
      \cline{1-1}\cline{3-5}
      \!\!contextual\!\! & & n/a & \!\!58.11\%\!\!  & \!\!56.98\% \!\!\\
      \cline{1-1}\cline{3-5}
      \!\!MIL (ours)\!\! & & \textbf{85.81\%} & \!\!\textbf{72.52\%}\!\! & \!\!\textbf{66.44\%}\!\! \\ 
      \hline
      \hline
      \!\!LSTM &\multirow{3}{*}{\!\!\rotatebox{90}{\!5-shot}\!\!} & 83.11\% & \!\!39.64\% \!\!& \!\!31.98\%\!\!\\
      \cline{1-1}\cline{3-5}
      \!\!contextual\!\!&  & n/a &\!\! 64.64\% \!\! & \!\!59.01\% \!\!\\ 
      \cline{1-1}\cline{3-5}
      \!\!MIL (ours) \!\! & & \textbf{88.75\%} & \!\!\textbf{78.15\%}\!\! & \!\!\textbf{70.50\%}\!\!\\
      \hline
    \end{tabular}
    \end{center}
    \vspace{-0.3cm}
    \caption{\footnotesize \!\! One-shot and 5-shot simulating pushing success rate with varying demonstration information provided at test-time. MIL can more successfully learn from a demonstration without actions and without robot state and actions than LSTM and contextual policies.
    }
    \label{table:pusher_ablation_video}
    \vspace{-0.4cm}
\end{wraptable}

The goal of our second set of experiments is to evaluate our approach on a challenging domain, involving 7-DoF torque control, a 3D environment, and substantially more physical and visual diversity across tasks.
The experiment consists of a family of simulated table-top pushing tasks, illustrated in Figure~\ref{fig:task}, where the goal is to push a particular object with random starting position to the red target amid one distractor. 
We designed the pushing environment starting from the OpenAI Gym PusherEnv, using the MuJoCo physics engine~\cite{gym,mujoco}.
We modified the environment to include two objects, vision policy input, and, across tasks, a wide range of object shapes, sizes, textures, frictions, and masses. We selected $116$ mesh shapes from \url{thingiverse.com}, $105$ meshes for meta-training and $11$ for evaluation. The meshes include models of chess pieces, models of animals like teddy bears and pufferfish, and other miscellaneous shapes. We randomly sampled textures from a set of over 5,000 images and used held-out textures for meta-testing. A selection of the objects and textures are shown in Figure~\ref{fig:objects}. For more experimental details, hyperparameters, and ablations, see Appendix~\ref{app:push}.

The performance of one-shot pushing with held-out objects, shown in Figure~\ref{fig:results-simreach}, indicates that MIL effectively learned to learn to push new objects, with $85.8\%$ one-shot learning success using the largest dataset size. Furthermore, MIL achieves, on average, $6.5\%$ higher success than the LSTM-based approach across dataset sizes. The contextual policy struggles, likely because the full demonstration trajectory information is informative for inferring the friction and mass of the target object.

In Table~\ref{table:pusher_ablation_video}, we provide two additional evaluations, using the largest dataset size. The first evaluates how each approach handles input demonstrations with less information, e.g. without actions and/or the robot arm state. For this, we trained each method to be able to handle such demonstrations, as discussed in Section~\ref{sec:no_action}. We see that the LSTM approach has difficulty learning without the expert actions. MIL also sees a drop in performance, but one that is less dramatic.
The second evaluation shows that all approaches can improve performance if five demonstrations are available, rather than one, despite all policies being trained for 1-shot learning. In this case, we averaged the predicted action over the 5 input demonstrations for the contextual and LSTM approaches, and averaged the gradient for MIL.


\begin{figure}[h]
\setlength{\unitlength}{0.5\columnwidth}
\begin{picture}(1.99,0.36) \linethickness{0.5pt}
\centering
\includegraphics[width=0.245\linewidth]{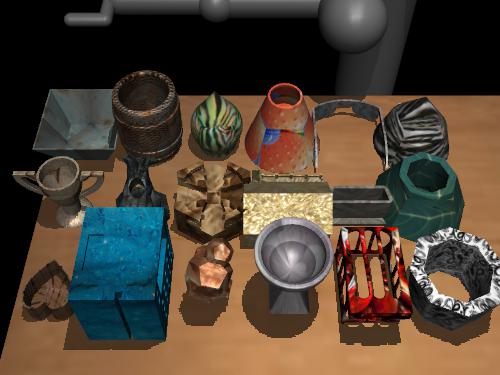}
\includegraphics[width=0.245\linewidth]{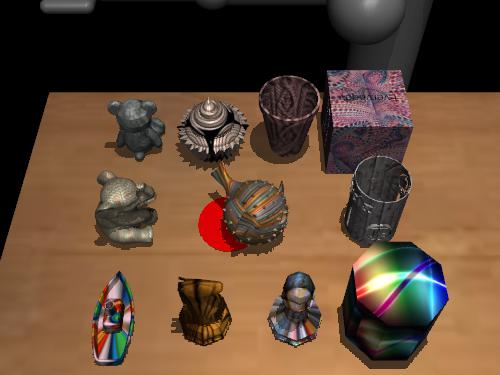}
\includegraphics[width=0.245\linewidth]{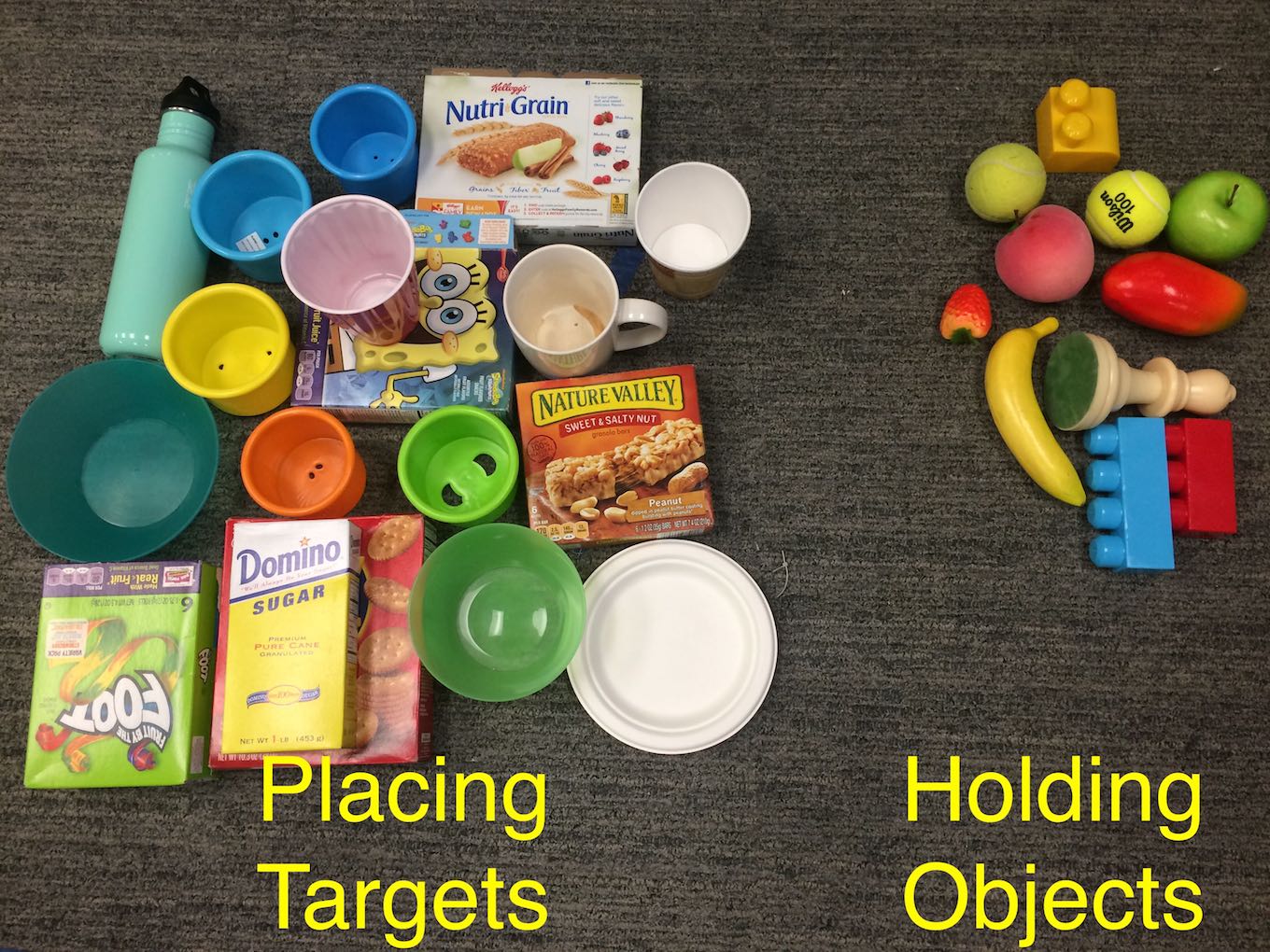}
\includegraphics[width=0.245\linewidth]{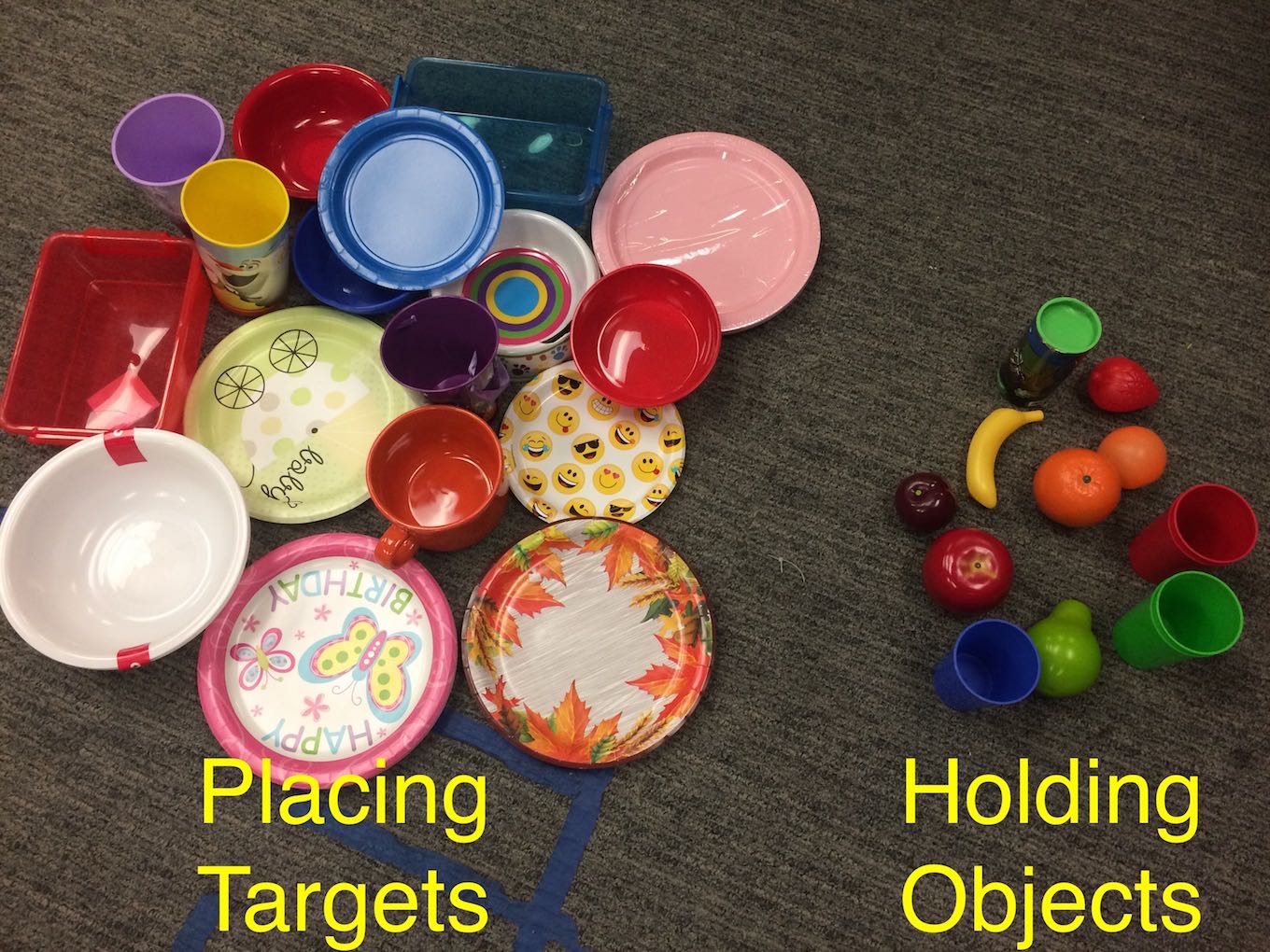}
\put(-2.0, -0.038){subset of training objects}
\put(-1.35, -0.038){test objects}
\put(-1.0, -0.038){subset of training objects}
\put(-0.35, -0.038){test objects}

\end{picture}
\vspace{0.1cm}
\caption{\footnotesize
Training and test objects used in our simulated pushing (left) and real-world placing (right) experiments. Note that we only show a subset of the $\sim\!\!100$ training objects used for the pushing and placing experiments, and a subset of the textures and object scales used for training and testing robot pushing.
\label{fig:objects}
\vspace{-0.5cm}
}
\end{figure}

\subsection{Real-World Placing}
\label{sec:placing}

The goal of our final experiment is to evaluate how well a real robot can learn to learn to interact with new, unknown objects from a single visual demonstration. Handling unseen objects is a challenge for both learning-based and non-learning-based manipulation methods, but is a necessity for robots to be capable of performing diverse tasks in unstructured real-world environments. In practice, most robot learning approaches have focused on much more narrow notions of generalization, such as a varied target object location~\cite{e2e} or block stacking order~\cite{rocky}.
With this goal in mind, we designed a robotic placing experiment using a 7-DoF PR2 robot arm and RGB camera, where the goal is to place a held item into
a target container, such as a cup, plate, or bowl, while ignoring two distractors.
We collected roughly $1300$ demonstrations for meta-training using a diverse range of objects, and evaluated one-shot learning using held-out, unseen objects (see Figure~\ref{fig:objects}). The policy is provided a single visual demonstration of placing the held item onto the target, but with varied positions of the target and distractors,
as illustrated in Figure~\ref{fig:task}. Demonstrations were collected using human teleoperation through a motion controller and virtual reality headset,
and each demo included
the camera video, the sequence of end-effector poses, and the sequence of actions -- the end-effector linear and angular velocities.
See Appendix~\ref{app:place} for more explanation of data collection, evaluation, and hyperparameters.

The results, in Table~\ref{table:real_placing_results}, show that the MIL policy can learn to localize the previously-unseen target object and successfully place the held item onto the target with $90\%$ success,
using only a
\begin{wraptable}{r}{5.5cm}
\vspace{-0.2cm}
    \begin{center}
    \vspace{-0.2cm}
    \begin{tabular}{l|c}
    method &  test performance\\
     \hline
      LSTM &  25\%   \\ 
      \hline
      contextual & 25\% \\
      \hline
      MIL & \textbf{90\%}\\
      \hline
      \hline
      MIL, video only & \textbf{68.33\%}
    \end{tabular}
    \end{center}
    \caption{\footnotesize One-shot success rate of placing a held item into the correct container, with a real PR2 robot, using $29$ held-out test objects. Meta-training used a dataset with $\sim\!\!100$ objects. MIL, using video only receives the only video part of the demonstration and not the arm trajectory or actions.}
    \vspace{-0.8cm}
    \label{table:real_placing_results}
\end{wraptable}
single visual demonstration with those objects. 
We found that the LSTM and contextual policies were unable to localize the correct target object, likely due to the modestly-sized meta-training dataset,
and instead placed onto an arbitrary
object, achieving $25\%$ success. Using the two-head approach described in~\ref{sec:no_action}, we also experimented with only providing the video of the demonstration, omitting the robot end-effector trajectory and controls. MIL can also learn to handle this setting, although with less success, suggesting the need for more data and/or further research.
We include videos of all placing policies in the supplementary video\footnote{For video results, see \url{https://sites.google.com/view/one-shot-imitation}}.

\section{Discussion and Future Work} 
\label{sec:conclusion}

We proposed a method for one-shot visual imitation learning that can learn to perform tasks using visual inputs from just a single demonstration. Our approach extends gradient-based meta-learning to the imitation learning setting, and our experimental evaluation demonstrates that it substantially outperforms a prior one-shot imitation learning method based on recurrent neural networks. The use of gradient-based meta-learning makes our approach more efficient in terms of the number of demonstrations needed during meta-training, and this efficiency makes it possible for us to also evaluate the method using raw pixel inputs and on a real robotic system.

The use of meta-imitation learning can substantially improve the efficiency of robotic learning methods without sacrificing the flexibility and generality of end-to-end training, which is especially valuable for learning skills with complex sensory inputs such as images. While our experimental evaluation uses tasks with limited diversity (other than object diversity), we expect the capabilities of our method to increase substantially as it is provided with increasingly more diverse demonstrations for meta-training. Since meta-learning algorithms can incorporate demonstration data from all available tasks, they provide a natural avenue for utilizing large datasets in a robotic learning context, making it possible for robots to not only learn more skills as they are acquire more demonstrations, but to actually become \emph{faster} and \emph{more effective} at learning new skills through the process.

\section*{Acknowledgments}
This work was supported by the National Science Foundation through IIS-1651843, IIS-1614653, IIS-1637443, and a graduate research fellowship, by Berkeley DeepDrive, by the ONR PECASE award N000141612723, as well as an ONR Young Investigator Program award. The authors would like to thank Yan Duan for providing a reference implementation of~\cite{rocky} and the anonymous reviewers for providing feedback.

\bibliography{references}

\newpage
\appendix

\section{Additional Experimental Details}

In this section, we provide additional experimental details for all experiments, including information regarding data collection, evaluation, and training hyperparameters.

\subsection{Simulated Reaching}
\label{app:reach}

\paragraph{Experimental Setup:}

In both vision and no-vision cases of this experiment, the input to the policy includes the arm joint angles and the end-effector position. In the vision variant, the $80\times64$ RGB image is also provided as input. In the non-vision version, the 2D positions of the objects are fed into the policy, but the index of the target object within the state vector is not known and must be inferred from the demonstration.
The policy output corresponds to torques applied to the two joints of the arm. A policy roll-out is considered a success if it comes within $0.05$ meters of the goal within the last $10$ timesteps, where the size of the arena is $0.6 \times 0.6$ meters.

To obtain the expert policies for this task, we use iLQG trajectory optimization to generate solutions for each task (using knowledge of the goal), and then collect several demonstrations per task from the resulting policy with injected Gaussian noise. At meta-test time, we evaluate the policy on $150$ tasks and $10$ different trials per task ($1500$ total trials) where each task corresponds to a held-out color. Note that the demonstration provided at meta-test time usually involves different target and distractor positions than its corresponding test trial. Thus, the one-shot learned policy must learn to localize the target using the demonstration and generalize to new positions, while meta-training must learn to handle different colors.

\paragraph{Hyperparameters:}

For all vision-based policies, we use a convolutional neural network policy with $3$ convolution layers each with $40$ $3\times3$ filters, followed by $4$ fully-connected layers with hidden dimension $200$. For this domain only, we simply flattened the final convolutional map rather than transforming it into spatial feature points. The recurrent policies additionally use an LSTM with $512$ units that takes as input the features from the final layer.
For non-vision policies, we use the same architecture without the convolutional layers, replacing the output of the convolutional layers with the state input.
All methods are trained using a meta batch-size of $5$ tasks. The policy trained with meta-imitation learning uses $1$ meta-gradient update with step size $0.001$ and bias transformation with dimension $10$. We also find it helpful to clip the meta-gradient to lie in the interval $[-20, 20]$ before applying it. We use the normal single-head architecture for MIL as shown in Figure~\ref{fig:latenttaskvar}.

\subsection{Simulated Pushing}
\label{app:push}

\paragraph{Experimental Setup:}

The policy input consists of a $125\times125$ RGB image and the robot joint angles, joint velocities, and end-effector pose. A push is considered a success if the center of the target object lands on the red target circle for at least 10 timesteps within a 100-timestep episode. The reported pushing success rates are computed over 74 tasks with 6 trials per task (444 total trials).

We acquired a separate demonstration policy for each task using the trust-region policy optimization (TRPO) algorithm. The expert policy inputs included the target and distractor object poses rather than vision input. To encourage the expert policies to take similar strategies, we first trained a single policy on a single task, and then initialized the parameters of all of the other policies with those from the first policy. When initializing the policy parameters, we increased the variance of the converged policy to ensure appropriate exploration.

\paragraph{Hyperparameters:}
For all methods, we use a neural network policy with $4$ strided convolution layers with $16$ $5\times5$ filters, followed by a spatial softmax and $3$ fully-connected layers with hidden dimension $200$. For optimization each method, use a meta-batch size of $15$ tasks. MIL uses $1$ inner gradient descent step with step size $\alpha = 0.01$, inner gradient clipping within the range $[-10, 10]$, and bias transformation with dimension $20$. The LSTM policy uses $512$ hidden units.

Because this domain is significantly more challenging than the simulating reaching domain, we found it important to use the two-head architecture described in section~\ref{sec:no_action}. We include an ablation of the two-head architecture in Table~\ref{table:pusher_ablation_head}, demonstrating the benefit of this choice.

\begin{table}[h]
    \begin{center}
    \begin{tabular}{l|c|c}
    method & 1-head & 2-head\\
     \hline
      MIL with 1-shot &  80.63\%  & 85.81\%   \\ 
      \hline
      MIL with 5-shot & 82.63\%  & 88.75\% \\
    \end{tabular}
    \end{center}
    \caption{Ablation test on 1-head and 2-head architecture for simulated pushing as shown in Figure~\ref{fig:latenttaskvar}, using a dataset with $9200$ demonstrations for meta-learning. Using two heads leads to significantly better performance in this domain.}
    \label{table:pusher_ablation_head}
\end{table}

\subsection{Real-World Placing}
\label{app:place}

\paragraph{Experimental Setup:}

The videos in the demo are composed of a sequence of $320\times 240$ RGB images from the robot camera. We pre-process the demonstrations by downsampling the images  by a factor of $2$ and cropping them to be of size $100\times 90$. Since the videos we collected are of variable length, we subsample the videos such that they all have fixed time horizon $30$. 

To collect demonstration data for one task, we randomly select one holding object and three placing containers from our training set of objects (see the third image of Figure~\ref{fig:objects}), and place those three objects in front of the robot in $8$ random
positions.  In this way, we collect $1293$ demonstrations, where we use $96$ of them as validation set and the rest as the training set. 

During policy evaluation, we evaluate the policy with $18$ tasks and $4$ trials per task ($72$ total trials) where we use $1$ placing target and $1$ holding object from the test set for each task. In addition, we manually code an "open gripper" action at the end of the trajectory, which causes the robot to drop the holding object. We define success as whether or not the held object landed in or on the target container after the gripper is opened.

\paragraph{Hyperparameters:}

We use a neural network policy with $3$ strided convolution layers and $2$ non-strided convolutions layers with $64$ $3\times3$ filters, followed by a spatial softmax and $3$ fully-connected layers with hidden dimension $100$. We initialize the first convolution layer from VGG-$19$ and keep it fixed during meta-training. We add an auxiliary loss besides our imitation objective, which regresses from the learned features at the first time step to the 2D positions of the target container. Additionally, we also feed the predicted 2D position of the target into the fully-connected network. MIL uses a meta-batch size of $12$ tasks, $5$ inner gradient descent steps with step size $0.005$, inner gradient clipping within the range $[-30, 30]$, and bias transformation with dimension $20$. We also use the single-head architecture for MIL just as what we do for simulated reaching. The LSTM policy uses 512 hidden units.

\end{document}